\documentclass[letterpaper, 10 pt, conference]{ieeeconf}  

\IEEEoverridecommandlockouts                              

\usepackage{amsmath,amsfonts}
\usepackage{algorithmic}
\usepackage{algorithm}
\usepackage{array}
\usepackage{textcomp}
\usepackage{stfloats}
\usepackage{url}
\usepackage{verbatim}
\usepackage{graphicx}
\usepackage{hyperref}
\usepackage{cite}
\usepackage{multirow}%
\usepackage{xcolor}%
\usepackage{booktabs}%
\usepackage{subfigure}

\title{\LARGE \bf
SIL: Symbiotic Interactive Learning for Language-Conditioned Human-Agent Co-Adaptation 
}

\author{Linus Nwankwo$^{*}$; Björn Ellensohn; Christian Rauch; Elmar Rueckert%
\thanks{This work is supported by the ``MINEVIEW'' project (\#FO999927835), funded by the Rep. of Austria, Fed. Min. of Env., Innovation and Tech.}%
\thanks{The authors are with the CPS, Technical University of Leoben, Austria.}%
\thanks{$^{*}$Corresponding author: {\tt\small linus.nwankwo@unileoben.ac.at}}%
\thanks{\copyright~2026 IEEE. 
Accepted for publication at the 35th IEEE Int.\ Conf.\ on Robot and Human Interactive Communication (RO-MAN 2026).}%
}

\begin{document}

\maketitle
\thispagestyle{empty}
\pagestyle{empty}

\begin{abstract}
    Today's autonomous agents, largely driven by foundation models (FMs), can understand natural language instructions and solve long-horizon tasks with human-like reasoning. However, current human-robot interaction frameworks largely follow a one-way master–apprentice technique where the embodied agent passively executes commands without reciprocal learning. This neglects the co-adaptive, multi-turn nature of everyday human-to-human interactions. We introduce symbiotic interactive learning (SIL), a bidirectional co-adaptation framework in a shared latent task space, where both the human and the agent maintain joint belief states that evolve with the interaction history. This enables proactive clarification, adaptive suggestions, and shared plan refinement. SIL leverages FMs for spatial perception and reasoning, together with a triplet-loss-trained neural encoder that grounds the FMs' outputs into task-specific latent representations. To support long-term stability as tasks evolve, SIL utilises episodic and semantic memory architectures, regularised via elastic weight consolidation to mitigate catastrophic forgetting. We evaluate SIL on simulated and real-world embodied tasks, including instruction following, information retrieval, query-oriented reasoning, and interactive dialogue, achieving a $90.4\%$ task completion rate and a belief alignment score of $\rho \approx 0.83$, an absolute improvement of about $20$ percentage points over the best ablations. Demos and resources: \url{https://linusnep.github.io/SIL/}.
\end{abstract}

\section{Introduction}\label{sec:introduction}
   The evolution of human-robot interaction (HRI) has reached a critical juncture, where the traditional one-way command-and-control-based approaches are no longer adequate for addressing complex, real-world tasks. Specifically, the state-of-the-art (SoTA) natural language-conditioned HRI frameworks~\cite{brohan2023can,nwankwo2025reli,lynch2023interactive} predominantly model communication as a unidirectional process. Humans issue commands, and the embodied agents attempt to interpret and execute them. This interaction pattern depicts a master-apprentice model, in which knowledge flows unidirectionally from the experienced master (human) to the learning apprentice (embodied agent), with the apprentice expected to absorb and apply the master's instruction without questioning or contributing novel insights back to the master. In other words, the agent remains purely a one‑way learner. 
   
\begin{figure}
    \centering
    \includegraphics[width=0.95\linewidth]{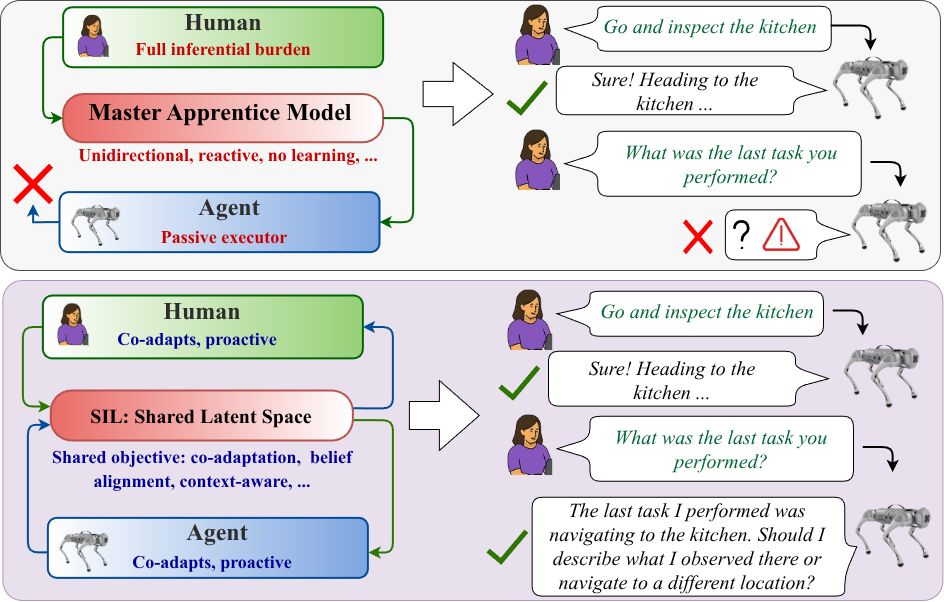}
    \caption{The unidirectional master apprentice model (top) places the entire inferential burden on the user (e.g., context, memory), requiring precise and unambiguous instructions for passive execution. In contrast, SIL (bottom) enables co-adaptive interaction, in which both actors iteratively update their shared latent beliefs to reduce ambiguity and inferential load.}
    \label{fig:sil-examp}
    \vspace*{-1.10\baselineskip}
\end{figure}

   Figure~\ref{fig:sil-examp} illustrates this one-way interaction mechanism. Although these methods~\cite{brohan2023can,nwankwo2025reli, lynch2023interactive} are effective for structured, short-term tasks, they do not capture the dynamic, reciprocal, and co-adaptive nature of human-to-human communication. In principle, they lack the mechanisms to represent, track, and align the evolving beliefs of both partners. As a result, interactions remain fragile to linguistic and contextual ambiguity, and therefore unsuited for the long-term adaptation required for robots to learn individual user preferences. The inferential burden rests largely on the human to compensate for the agent's static understanding, preventing the natural and efficient collaboration common in human teams.

       \begin{figure*}
        \centering
        \includegraphics[width=0.942\linewidth]{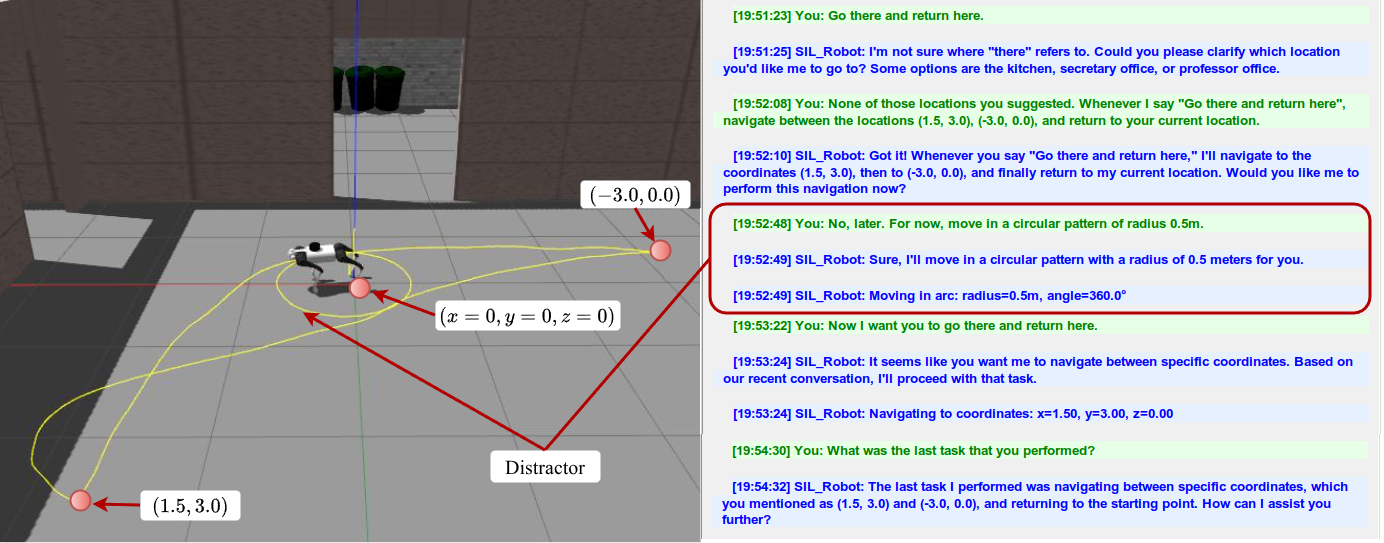}
        \caption{An example of SIL's contextual dialogue-based grounding: upon receiving an ambiguous instruction, a clarification dialogue was triggered. The agent offers candidate interpretations based on prior interactions, resolves the intent, and executes the navigation task (yellow path), even after a distractor.}
        \label{fig:TradVsSIL}
        \vspace*{-1.10\baselineskip}
    \end{figure*}
    Motivated by these challenges, we propose a symbiotic interactive learning (SIL) framework that reframes language-conditioned HRI as a dynamic, co-adaptive process. Rather than treating humans as a fixed command source and agents as passive executors, SIL models both parties as adaptive systems that maintain and align their beliefs within a shared latent task space. This pushes beyond the passive action execution to active collaboration. For instance, given the ambiguous instruction shown in Figure~\ref{fig:TradVsSIL} (\textit{``go there and return here''}), the agent not only seeks clarification, but also leverages shared-context derived from prior interactions to proactively suggest likely interpretations (e.g., \textit{`Some options are: $\cdots$'}). Additionally, the agent contributed its observations and retained memory of past interactions (e.g., \textit{``The last task I performed was $\cdots$''}). This represents a crucial shift from the traditional methods toward interaction grounded in mutual understanding through dialogue. Accordingly, the primary contributions of this work are as follows:
\begin{itemize}
        \item \textbf{Characterisation of unidirectional learning problem in language-conditioned HRI.} We analyse the limitations of the master-apprentice model, common with the SoTA language-conditioned HRI methods and introduce SIL, a bidirectional symbiotic framework that enables continuous, mutual adaptation between human and agent within a shared latent task space.
        \item \textbf{Shared belief representation and alignment.} We introduce a mechanism to explicitly represent, measure, and align human and agent beliefs through a shared latent representation, to enable targeted clarification, proactive suggestions, and quantifying understanding. 
        \item \textbf{Continual learning with structured memory.} We integrate a continual learning architecture with structured episodic and semantic memory to preserve knowledge across interactions, and mitigate catastrophic forgetting~\cite{kirkpatrick2017overcoming} of learned task representation.
        \item \textbf{Extensive empirical evaluation.} We evaluate SIL across diverse task domains in both real-world and simulated environments, with the results showing significant improvements in interaction efficiency and robustness compared with the static unidirectional baselines.
\end{itemize}

\section{Related Works}\label{sec:relatedworks}
\subsection{Foundation Models for Language-Conditioned HRI}
    Foundation models (FMs)~\cite{hurst2024gpt,team2023gemini, liu2024deepseek, radford2021learning} have substantially reshaped language-conditioned HRI from the traditional, rigid symbolic parsing towards knowledge-guided embodiment, in which autonomous agents can interpret unconstrained natural language instructions grounded in rich perceptual context. Frameworks such as \textit{SayCan}~\cite{brohan2023can}, \textit{Interactive Language}~\cite{lynch2023interactive}, \textit{ProgPrompt}~\cite{singh2023progprompt}, and \textit{TCC}~\cite{nwankwo2024conversation}, among others, demonstrate that FMs can be effectively grounded to support task execution from free-form natural language instructions.

    However, in these frameworks, language understanding is typically treated as a front-end module, often decoupled from the agent's core reasoning and planning. This separation results in a largely unidirectional interaction paradigm where dialogue and action remain disjoint, leaving the agent as a reactive executor of user commands. Adaptation likewise flows one way: only the agent adjusts to human input, typically encoded as structured reasoning. 
    We argue that scaling language-conditioned HRI frameworks with FMs alone is insufficient; the interaction mechanism itself must evolve so that language becomes a medium for shared reasoning rather than merely a one-way command transmission.
 
\subsection{Symbiotic Human-Robot Interaction and Current Gaps}
    Long-term interaction requires a shared understanding that evolves through mutual adaptation~\cite{sciutti2018humanizing}.
    Prior works have explored mutual adaptation in contexts such as shared autonomy~\cite{javdani2015shared}, collaborative planning~\cite{7451736}, and predefined state-action interaction models~\cite{rosenthal2010effective}. Accordingly, Javdani et al.~\cite{javdani2015shared} infer user intent for shared control via hindsight optimisation, but assumed a fixed human policy without modelling evolving human beliefs. Nikolaidis et al.~\cite{7451736} formalise mutual adaptation through bounded-memory that captures policy convergence over discrete action spaces; however, their approach does not support continuous latent belief alignment or natural language interaction. Rosenthal et al.~\cite{rosenthal2010effective} demonstrate symbiotic behaviour through structured help-seeking strategies, but rely on hand-crafted state representations rather than learned belief embeddings.

    More recently, learning-based methods adapt agent policies to human preferences~\cite{mahadevan2024generative,christiano2017deep}, and dialogue-based approaches~\cite{liu2018dialogue,dong2025toward,nwankwo2024multimodal} to support interactive instruction. However, they generally lack mechanisms for continuous, bidirectional co-adaptation of internal beliefs and decision-making, an essential ingredient for robust, collaborative partnerships.

    Overall, in the current language-conditioned HRI frameworks, three interrelated gaps exist: (i) predominance of unidirectional adaptation, where only the agent adapts to a static human model; (ii) modular separation between language understanding, learning, and belief modelling, rather than a unified inferential process; and (iii) lack of mechanisms for sustained, bidirectional belief alignment through natural language interaction.
    SIL addresses these gaps directly. 

\section{Method}\label{sec:method}
\begin{figure*}
        \centering
        \includegraphics[width=0.88\linewidth]{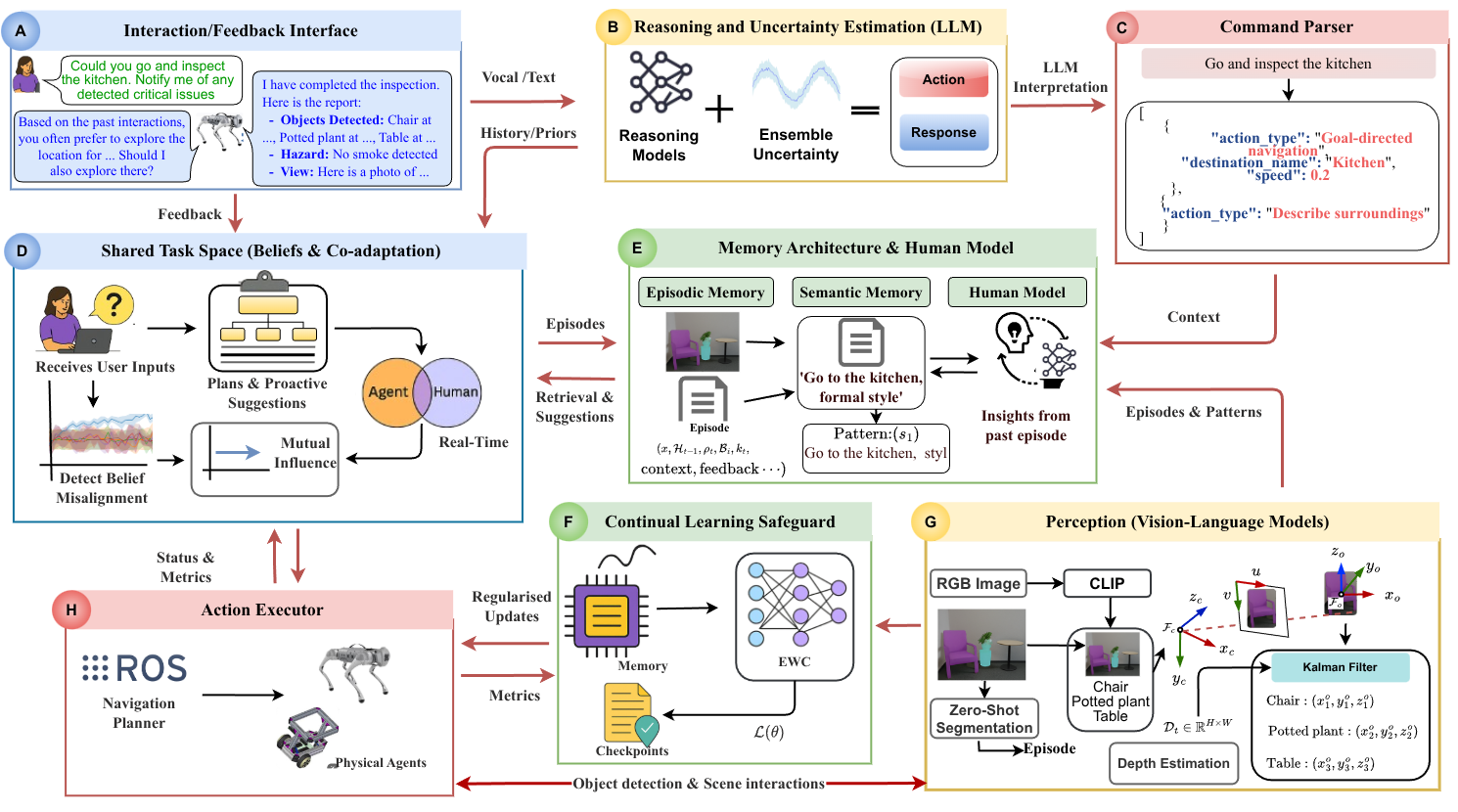}
        \caption{Overview of SIL's architecture. Human instructions are received through the natural-language interaction interface (A) and passed to the LLM ensemble for intent parsing (B \& C). Internally, the agent maintains belief states in a shared latent task space. This is updated through co-adaptation dynamics and aligned via cosine similarity (D, E, \& F). Visual grounding is achieved through pre-trained vision–language models that segment and project objects into 3D coordinates (G). Action plans are executed through the action executor (H) while providing feedback in the form of progress updates, error reporting, and adaptive suggestions. The memory architecture ensures continual adaptation over time.}
        \label{fig:sil-framework}
        \vspace*{-1.10\baselineskip}
\end{figure*}

   We address the problem of unidirectional grounding common with recent natural-language-conditioned human-robot interaction (HRI) frameworks. We postulate that effective HRI requires continuous mutual co-adaptation that mirrors human-to-human communication. This section presents the formal details of our proposed framework. Fig.~\ref{fig:sil-framework} shows the architectural overview of our approach.

\subsection{Problem Description: Unidirectional Adaptation}\label{sec:prob-def}
    Recent language-conditioned HRI frameworks~\cite{brohan2023can,lynch2023interactive,singh2023progprompt,zhou2024navgpt,nwankwo2024conversation} commonly adopt a unidirectional grounding architecture, in which natural language instructions $x \in \mathcal{X}$ and contextual information $c \in \mathcal{C}$ directly map to robot executable actions $y \in \mathcal{Y}$: $y_t \sim \pi_{\theta}(y \mid x_t \,,\,c_t)$, where $\pi_\theta$ denotes a conditional policy parameterised by the pre-trained weights $\theta$  that often remain fixed during deployment. Under this formulation, the agent’s interpretation of language and context is assumed to be fully encoded in $\theta$. Consequently, the agent maintains a time-invariant latent belief state $\mathcal{B}^A_\text{static}=\{\mathbf{z}^A_\text{static}\},\text{with}\,\tfrac{\partial \theta}{\partial t}=0$, where $\mathbf{z}^A_{\text{static}} \in \mathcal{Z} \subseteq \mathbb{R}^d$ represents the agent’s static latent representation. 

    In contrast, the human's belief state $\mathcal{B}^H_t$ evolves dynamically as the user observes and adapts to the agent’s behaviour. Because the agent neither observes nor explicitly models $\mathcal{B}_t^H$, the alignment depends solely on the human to adapt to the agent's rigid interpretation manifold. This asymmetric adaptation mechanism leaves $\mathbf{z}^A_{\text{static}}$ and the human's evolving belief $\mathbf{z}_t^H$ persistently misaligned, and thereby precludes clarification, mutual disambiguation, and preference modelling. Our objective, therefore, is to overcome these limitations through bidirectional co-adaptation.
    
\subsection{SIL: Belief Representation and Co-Adaptation}\label{sec:sil}
    We reconceptualise the problem (Section~\ref{sec:prob-def}) through co-evolving belief states within a shared latent task space $\mathcal{Z} \subseteq \mathbb{R}^d$. Each actor $i \in \{H, A\}$ maintains a structured belief state $\mathcal{B}_t^i$, which evolves based on ongoing interactions, and is formally characterised by the tuple: 
\begin{equation}\label{eq:belief}
            \mathcal{B}^i_t=\left(\mathbf{z}^i_t\,,\,\mathbf{k}^i_t\,,\,\mathbf{u}^i_t\,,\,\mathcal{H}^i_t\,,\, \Psi^i_t\right)\,,
\end{equation}
    where $\mathbf{z}_t^i \in \mathcal{Z}$ is the latent task embedding that encodes goal or intent understanding;  $\mathbf{k}_t^i \in [0, 1]$ is a confidence scalar that measures the belief certainty; $\mathbf{u}_t^i$ is an uncertainty representation; and the temporal memory buffer $\mathcal{H}_t^i$ is a bounded history of interaction embeddings supporting sequential reasoning. The auxiliary model $\Psi^i_t$ encodes the actor-specific parameters; for the human ($i=H$), this is a preference model $\mathbf{p}_t$ that captures personalised goals, styles, or feedback tendencies, while for the agent ($i=A$), it maintains a running estimate of the human's latent embedding and confidence, updated via Eq.~\eqref{eq:belief-change}, where $\mathcal{B}_t^H$ is the agent's inferred estimate of the human's state.
    Unlike the traditional approach, these belief states (Eq.~\eqref{eq:belief}) are co-evolved: the agent not only updates its internal representation based on the observed inputs and feedback, but also reasons over its estimate of the human’s latent state, and vice versa.
     The belief update is governed by a mutual recursive inference, defined by the transition: $\mathcal{B}^i_{t +1} \sim \Phi^i(\mathcal{B}^i_t\,,\,\mathcal{O}^i_t\,,\,\hat{\mathcal{B}}_t^{-i|i})$, where $\Phi^i$ is a belief transition operator, $\mathcal{O}^i_t$ are local observations, and $\hat{\mathcal{B}}_t^{-i|i}$ is the cross-agent belief. 
     
    To operationalise this, we define a bidirectional cross-agent influence mechanism where each actor’s belief is modulated by the other’s. We compute the influence vectors using learned transformations over the other’s latent embedding as:
\begin{equation}\label{eq:lin-trans}
    \delta_\text{infl}^A = \tanh(W_{HA}\;\mathbf{z}^H_t), \quad \delta_\text{infl}^H = \tanh(W_{AH}\;\mathbf{z}^A_t),
\end{equation}
    where $W_{HA} \in \mathbb{R}^{d \times d}$ and $W_{AH} \in \mathbb{R}^{d \times d}$ are weight matrices that capture human-to-agent and agent-to-human influence dynamics, respectively. Both matrices are initialised with small random values $(\sigma = 0.01)$ and updated online based on accumulated interaction gradients. Therefore, given a latent embedding of the most recent interaction $\mathbf{z}_\text{new}$, and observed interaction success $s_t \in [0, 1]$, and a set of tuning coefficients $\eta_i \geq 0$, we update the task embeddings for both actors as: 
\begin{equation}\label{eq:belief-change}
        \begin{split}
            \mathbf{z}^{A}_{t+1} = \eta_{1} \mathbf{z}^{A}_{t} 
            + \eta_{2} \mathbf{z}^{A}_{\text{new}} 
            + \eta_{3} \left( \alpha_{A} \cdot s_{t} \cdot \delta^{A}_{\text{infl}} \right)\\
            \mathbf{z}^{H}_{t+1} = \eta_{4} \mathbf{z}^{H}_{t} 
             + \eta_{5} \mathbf{z}^{H}_{\text{new}} 
            + \eta_{6} \left( \alpha_{H} \cdot (2 - s_{t}) \cdot \delta^{H}_{\text{infl}} \right),
        \end{split}
\end{equation}
    where $\alpha_A \,,\,\alpha_H \in [0, 1]$ are adaptation rates for the agent and human, respectively.
    Notably, the $(2 - s_t)$ factor ensures that the agent execution failures ($s_t \approx 0$) provide a strong signal for the human to adapt (e.g., rephrasing or simplifying instructions). Conversely, when execution succeeds $(s_t \approx 1)$, the human influence reduces, reflecting that less corrective adjustment is needed. This asymmetric scaling encodes the intuition that failures are more informative signals for belief revision than successes. All latent vectors are $\ell_2$-normalised after each update for numerical stability.
    
    Furthermore, to monitor the interaction quality, we compute a confidence-weighted belief alignment $\rho_t$ based on the similarity between the human and agent task embeddings as:
    \begin{equation}\label{eq:belief-align}
    \rho_t = \left( \frac{1 + \cos({\mathbf{z}_t^H \,,\, \mathbf{z}_t^A})}{2} \right) \cdot \mathbf{k}^H_t \cdot \mathbf{k}^A_t \;,\; \rho_t \in [0, 1].
\end{equation}
    \textbf{Clarification protocol.} Intuitively, Eq.~\eqref{eq:belief-align} quantifies the confidence-weighted agreement between the human's and agent's task-level beliefs. If $\rho_t$ falls below the misalignment threshold $\tau_\text{mis} \in [0, 1]$, we initiate a clarification protocol to resolve discrepancies prior to further execution. This proceeds in three stages: (i) the agent identifies the sources of uncertainty by inspecting its uncertainty map $u^A_t$ (e.g., whether the ambiguity lies in intent, parameters, or destination), and generates candidate interpretations by sampling the LLM ensemble (Section~\ref{sec:llm-assemble}) and retrieving semantically similar past episodes from episodic memory (Section~\ref{sec:mem}); (ii) these candidates are ranked by their alignment with the current belief state and presented to the human as alternative options (e.g., ``Some options are: ...''); and (iii) the human's selection is used to update both $z^H_t$ and $z^A_t$ via Eq.~\eqref{eq:belief-change}, with the successful resolution stored as a positive episode in memory. This ensures proactive intervention in cases of latent misunderstanding, rather than reactive correction. 
    
    \textbf{Encoder architecture and training.} To support these dynamics, we train a lightweight neural encoder $\phi: \mathbb{R}^{768} \rightarrow \mathcal{Z}$ that maps linguistic inputs and dialogue history into latent task embeddings.
    Each utterance $x_j$ is first encoded by a frozen pre-trained sentence transformer into a contextual representation $u_j \in \mathbb{R}^{768}$. We then aggregate the dialogue history through attention pooling, $\mathcal{H}_t = \text{AttnPool}(\{u_j\}_{j=1}^J)$. The resulting representation $(x_t\,,\,\mathcal{H}_t)$ is projected by the $\phi$ into $\mathcal{Z}$. The encoder architecture and all associated hyperparameters are presented in the Appendix~\ref{sec:imp-details}.

    \textbf{Encoder initialisation and online updates.} At the start of interaction, the human belief state $\mathcal{B}^H_0$ is initialised by projecting the first user utterance into the latent space, with confidence $\mathbf{k}^H_0$. The agent's belief state $\mathcal{B}^A_0$ is initialised as a noisy copy of the human projection (additive Gaussian noise, $\sigma_{\text{init}}$) with $\mathbf{k}^A_0$, reflecting higher initial uncertainty about the human's intent. If prior interaction history exists, the encoder resumes from the most recent checkpoint.
    We continually update the encoder, $\phi$ (i.e., after every completed interaction episode), using a triplet contrastive loss objective $\mathcal{L}_3$ (Eq.~\eqref{eq:contrastive-loss-obj}) that organises the latent space based on semantic and behavioural similarity. After each interaction (anchor, $x_a$), we retrieve a positive sample $x_p$ (successful, semantically similar past command) and a negative example $x_n$ (semantically dissimilar or failed interaction) from the episodic memory (Section~\ref{sec:mem}). Our objective is to minimise:
    \begin{equation}\label{eq:contrastive-loss-obj}
        \mathcal{L}_{3} = \max(\mid\mid\phi(x_a) - \phi(x_p)\mid\mid^2_2 - \mid\mid\phi(x_a) - \phi(x_n)\mid\mid^2_2 + m, 0)
    \end{equation}
    where $m$ is a margin parameter. This objective encourages successful interactions to cluster in latent space, while pushing away failed or misaligned examples. To preserve long-term stability, we incorporate an elastic weight consolidation (EWC)~\cite{kirkpatrick2017overcoming} penalty, $\mathcal{L}_\text{ewc}$, which prevents the encoder from forgetting previously important representations. Thus, our total learning objective becomes: $\mathcal{L} = \mathcal{L}_3 + \mathcal{L}_{ewc}$, (Eq.~\eqref{eg:Tlos}).
      
\subsection{Memory and Continual Learning Safeguards}\label{sec:mem}
    \textbf{Memory.} To support long-term adaptation and personalisation, SIL employs a dual-component memory architecture (Fig.~\ref{fig:sil-framework}E) comprising structured episodic and semantic memory. These components jointly enable the agent to recall past experiences, generalise from them, and safeguard prior knowledge.
    The episodic memory functions as a fixed-size buffer, $\mathcal{M}_{\text{ep}}$, that stores interaction records. Each episode $e_i$ records the raw user input, agent response, execution context, latent representation $\phi(x_i)$, internal belief states $(\mathcal{B}^H_i\,,\, \mathcal{B}^A_i)$, belief alignment score $\rho_i$, success signal $s_i$, and timestamp.
    Semantic memory, on the other hand, consolidates accumulated interactions into generalised patterns. It distils episodic experiences into abstract knowledge organised by task type, including success patterns, failure patterns, common clarification triggers, and co-adaptation patterns that capture periods of sustained convergence of beliefs.

    \textbf{Human model \& memory retrieval.} Alongside the memory components, SIL maintains a lightweight human model (Fig.~\ref{fig:sil-framework}E) that tracks user communication style and preferences. This model learns incrementally from each interaction using exponential moving averages over features such as verbosity, formality, and specificity.
    The memory retrieval is belief-aware. Given a new command $x_t$, and a candidate past episode $e_i$, we compute a relevance score $\mathcal{S} (x_t ,e_i)$ that combines semantic similarity and belief alignment:
\begin{equation}
        \mathcal{S}(x_t, e_i) =  w_\text{s}\mathcal{S}_{s}(\phi(x_t), \phi(x_i)) + w_{b}\mathcal{S}_{b}(\mathcal{B}^A_t, \mathcal{B}_i),
\end{equation}
    where $\mathcal{S}_s$ measures sentence-level similarity, and $\mathcal{S}_b$ compares the current agent's belief $\mathcal{B}_{t}^{A}$ with the stored belief state $\mathcal{B}_i$: $S_b\!\bigl(\mathcal{B}^A_t,\,\mathcal{B}_i\bigr) \,=\, \cos\!\bigl(\mathbf{z}^A_t,\,\phi(x_i)\bigr) \,\cdot\, \mathbf{k}^A_t \,\cdot\, \rho_i\,$. This ensures that retrieved episodes are not only semantically close but also consistent with the agent’s internal state. The weights $w_{s}$ and $w_{b}$ balance linguistic and belief-driven signals, and final retrieval probabilities are obtained through a softmax, i.e.,  $\pi (i \mid x_t) = \text{softmax}(\mathcal{S}(x_t, e_i)/\tau)$.

    \textbf{Continual learning safeguard.} While the memory architecture enables continual refinement, the online fine-tuning of the latent task encoder $\phi$ introduces the risk of catastrophic forgetting, whereby newly learned tasks overwrite previously acquired knowledge. To mitigate this, we employ the EWC mechanism~\cite{kirkpatrick2017overcoming} as a continual learning safeguard. We monitor interaction performance over a rolling window of recent episodes. A task shift is detected when the current success rate drops significantly below the rolling average. Upon detecting a shift, we checkpoint the current model parameters and trigger knowledge preservation mechanisms.

     Next, for each completed task $k$, we store the optimal parameters $\theta^{*(k)}$, and estimate the Fisher information matrix $\mathbf{F}^{(k)}$ by averaging squared gradients over recent interactions:
\begin{equation}
       \mathbf{F}_i^{(k)} = \frac{1}{N} \sum_{n=1}^{N} \left( \frac{\partial \mathcal{L}}{\partial \theta_i} (x_n) \right)^2.
\end{equation}
    This matrix quantifies the relative importance of each parameter. During future updates, we impose an EWC regularisation penalty to resist changes to parameters deemed critical for prior tasks. The final total loss function thus becomes:
\begin{equation}\label{eg:Tlos}
    \mathcal{L}(\theta)
    \;=\;
    \mathcal{L}_3(\theta)
    \;+\;
    \underbrace{\frac{\lambda}{2}
    \sum_{k=1}^{K}\sum_{i}
    \mathbf{F}^{(k)}_i
    \left(\theta_i - \theta^{*(k)}_i\right)^{2}}_{\mathcal{L}_\text{ewc}(\theta)}
\end{equation}
    where $\mathcal{L}_3$ is the triplet contrastive loss (Eq.~\eqref{eq:contrastive-loss-obj}), and $\lambda$ is an importance coefficient that balances plasticity and stability. 
    
\subsection{Uncertainty-Aware Language Understanding and Parsing}\label{sec:llm-assemble}
    To ensure robust intent recognition and prevent unsafe execution of ambiguous instructions, SIL combines ensemble-based reasoning~\cite{chipman2006bayesian} with linguistic feature analysis and context-aware parsing for reliable interpretation of users' command inputs (Fig.~\ref{fig:sil-framework}B \& C). 
    For any user command $x$, we generate $K$ distinct interpretations by sampling the distribution from LLM at varying temperatures, $\mathcal{T} = \{\mathcal{T}_1, \mathcal{T}_2, \dots, \mathcal{T}_K\}$. Each sample yields a candidate interpretation $y_k$, forming the ensemble $\mathcal{Y} = \{y_1, y_2, \dots, y_K\}$. Concretely, every candidate $y_k$ is encoded by a frozen pre-trained sentence transformer into a dense vector $\mathbf{v}_k\!\in\!\mathbb{R}^{d_v}$. To estimate the dispersion within this ensemble, we compute the average pairwise cosine distance across all ensemble as:
\begin{equation}\label{eq:dispersion}
    \mathbf{D}(x)
      \;=\;
      \frac{2}{K(K-1)}
      \sum_{i<j}
      \bigl(1 - \cos(\mathbf{v}_i,\,\mathbf{v}_j)\bigr),
\end{equation}
    where $\cos(\cdot,\cdot)$ denotes cosine similarity. High dispersion indicates that the ensemble members diverge semantically, signalling uncertainty about the correct interpretation.
    Concurrently, we extract linguistic confidence features $\mathbf{C_\text{ling}}(x) \in [0,1]$ from each ensemble response using a rule-based classifier that detects hedging expressions, parameter specificity, semantic completeness, and structural complexity. Therefore, we compute the overall uncertainty metric $\mathbf{U}(x)$ as:
\begin{equation}
        \mathbf{U}(x) = \alpha_u \mathbf{D}{(x)} + (1-\alpha_u) (1 - \mathbf{C}_\text{ling}(x)) + \beta_u \mathbf{C}_\text{ctx}(x),
\end{equation}
    where $\alpha_u\,,\, \beta_u$ are empirically determined weights, and $\mathbf{C}_\text{ctx}(x)$ quantifies contextual novelty by measuring the maximum cosine similarity between the current input embedding $\phi(x)$ and all episodes in the memory buffer: $\mathbf{C}_\text{ctx}(x) = 1 - \max_i\mathcal{S}_s(\phi(x)\,,\, \phi(x_i))$. High $\mathbf{U}(x)$ triggers the clarification protocol described in Section~\ref{sec:sil}.
    To derive the final, uncertainty-aware interpretation $\hat{y}$, we apply a weighted consensus mechanism over the ensemble: 
\begin{equation}\label{eq:llm-int}
        \hat{y} = \arg\max_{y \in \mathcal{Y}} \sum_{k=1}^{K} w_k\cdot K(y_k \,,\,y) (1 - \mathbf{U}_k(x)),
\end{equation}
    where $w_k$ are temperature-dependent sampling weights, that favour conservative responses,  $\mathbf{U}_k(x)\!=\!1-\mathbf{C}_{\mathrm{ling}}(y_k)$ denotes the per-member uncertainty, and $K(\cdot,\cdot)\!\in\![0,1]$ is a cosine-similarity kernel evaluated in the same sentence-embedding space used for $\mathbf{D}(x)$. Eq.~\eqref{eq:llm-int} ensures that interpretations with lower uncertainty and greater consensus contribute more to the final decision, while high-uncertainty interpretations are down-weighted.

    In terms of command parsing (Fig.~\ref{fig:sil-framework}C), we employ a hierarchical approach combining structured JSON with the resulting LLM-guided interpretation, Eq.~\eqref{eq:llm-int}, to transform the natural language commands into executable actions.

 \subsection{Multimodal Perception and Action Execution}\label{sec:action-execution}
    SIL's visuospatial and action execution pipelines (Fig. \ref{fig:sil-framework}G \& H) ground interactions in the physical reality. We employed the SAM~\cite{kirillov2023segment} to perform instance segmentation and CLIP~\cite{radford2021learning} for zero-shot open-vocabulary object classification via joint vision-language embeddings.

    We derive 3D object coordinates by projecting 2D mask centroids into 3D space using camera intrinsic and depth data, with monocular depth estimation~\cite{ranftl2020towards} to supplement unreliable depth. These coordinates are transformed into global frames using calibrated ROS~\cite{quigley2009ros} transformations to enable agents to interpret and execute spatial commands (e.g., ``go to the chair"). We utilised a Kalman filter to track objects over time and smooth pose estimates.
    For navigation (Fig.~\ref{fig:sil-framework}H), we rely on the ROS planning stack~\cite{macenski2020marathon2} for path planning, obstacle avoidance, and sensor-based information retrieval. We employ a Rao-Blackwellized algorithm~\cite{grisetti2007improved} and AMCL~\cite{thrun2001robust} to learn occupancy grid representations and localise the agent in the environment. With the agent localised, zero- and few-shot goal-directed navigation commands (e.g., `head to the kitchen') become interpretable.

\section{Experiments and Results}\label{sec:results}
    We empirically evaluate SIL across simulated and real-world environments. We focus on its co-adaptive mechanisms for belief alignment, memory, and preference learning across five key dimensions: (i) instruction execution under ambiguity and temporal complexity, (ii) long-term memory and retention, (iii) contextual reasoning, (iv) clarification and proactive dialogue, and (v) preference-based personalisation. 
       
    \subsection{Experiment Setup}
    We deployed SIL on two mobile platforms (Unitree Go1 and our customised Segway robot, see Fig.~\ref{fig:qualitative-visualisation}) equipped with an RGB-D camera and LiDAR.
    We utilised GPT-4o~\cite{hurst2024gpt} as the LLM backbone in all the experiments. Simulation was conducted in Gazebo with an Nvidia RTX-4090, and in the real world with a Lenovo ThinkBook i7. Further details and hyperparameters are provided in Appendix~\ref{sec:imp-details}.
     
    Similar to TCC~\cite{nwankwo2024conversation}, we conducted $350$ human-robot interaction episodes distributed across the five task domains described in Section~\ref{sec:task-domain} as: EIF ($n{=}120$), MIIR ($n{=}60$), QOR ($n{=}80$), PDS ($n{=}40$), and LPL ($n{=}50$), where $n$ is the interaction trajectories.
    Since no existing framework jointly addresses bidirectional belief co-adaptation, continual learning, and shared latent grounding in language-conditioned HRI, we therefore evaluate SIL against a static LLM (GPT-4o~\cite{hurst2024gpt} without memory or adaptation, representing the unidirectional master-apprentice model) and five ablated variants, each disabling one core component (Section~\ref{sec:ablation}).
    
\subsection{Task Domains and Dataset}\label{sec:task-domain}
    To ensure robust evaluation that captures the complexities and ambiguities across real-world interactions, we designed task instructions that test SIL on the following capabilities:
\paragraph{Embodied Instruction Following (EIF)}\label{sec:eif}
    This includes single-turn direct instructions (e.g., ``move forward $1.5~m$"), multi-turn long-horizon tasks that require sequential actions and context retention (e.g., ``go to the professor's office, describe the objects you can see, and then return to the starting point"), and constraint-rich tasks that involve conditional reasoning, such as ``navigate between the coordinates $(2,3,0)$ and $(-3, 2,0)$, if the round-trip time at max speed is under $15~s$, otherwise, rotate in place and report orientation". 
\paragraph{Memory-Based Interactive Information Retrieval (MIIR)}\label{sec:miir}
    This evaluates SIL’s memory architecture and anti-forgetting safeguards through two categories of queries: (i) retrospective queries, that require episodic recall and spatial reasoning (e.g., `what was the last location you visited?', and, queries probing recall after distractor tasks such as 'was there a chair in the last visited area?'), and (ii) procedural queries, which test the stability of learned command aliases. For example, we taught the agent that `patrol now' implies `navigate between the corridor and the kitchen', then issued several distractor tasks, before reissuing `patrol now' to test whether EWC preserved the newly taught behaviour.
\paragraph{Query-Oriented Reasoning (QOR)}\label{sec:qor}
    This focused on tasks that probe deductive, hypothetical, and inductive inference. Deductive tasks required logical reasoning over the known spatial map, e.g., `how long would it take you to get to the kitchen?' Hypothetical tasks test the agent’s ability to reason over its internal world model without execution, e.g., `If you were in the kitchen, which locations are directly visible?' Inductive tasks assessed generalisation from experience, e.g., `Based on the offices you have observed, what object is often found in them?' Collectively, these tasks test SIL’s capacity to handle structured reasoning across spatial knowledge and counterfactual scenarios. 
\paragraph{Proactive Dialogue and Suggestion (PDS)}\label{sec:pds}
    Here, we issued ambiguous instructions (e.g., `head to the location and return here') and evaluated whether SIL requested clarification, inferred intent from history, or proposed suitable alternatives. We also assessed the contextual appropriateness of proactive suggestions with four independent human raters (Avg. age $32\pm3$, males) who scored each suggestion on a 3-point scale (irrelevant / partially relevant / fully relevant). 
\paragraph{Long-Term Preference Learning (LPL)}\label{sec:ltpl}
    In extended multi-turn sessions, we measured SIL’s adaptation to user communication styles and preferences. For example, we issued instructions such as `from now on, when I say quick, I mean move at your fastest speed'. We then evaluated SIL on whether it retained and applied these preferences in later commands after distraction tasks.
    
\subsection{Evaluation Metrics}
    We quantitatively evaluate SIL with the following metrics:  
    (i) \textbf{Task Completion Rate (TCR)}: This represents the proportion of tasks correctly executed.  A task is considered successful if the agent reaches the correct goal state (e.g., arriving within $0.5m$ of the target location for navigation, or producing a factually correct response for reasoning tasks) without triggering a fatal error (e.g., navigation collisions with obstacles or execution of an action that contradicts the user's stated intent (e.g., navigating to the wrong destination)).
    (ii) \textbf{Belief Alignment ($\rho$)}: This quantifies the confidence-weighted similarity between human and agent belief embeddings, as described in Eq.~\eqref{eq:belief-align}.  
    (iii) \textbf{Clarification Efficiency (CE)}: This measures the mean number of clarification requests per successful task. 

\subsection{Quantitative and Qualitative Results}
\begin{figure}
        \centering
        \includegraphics[width=0.90\linewidth]{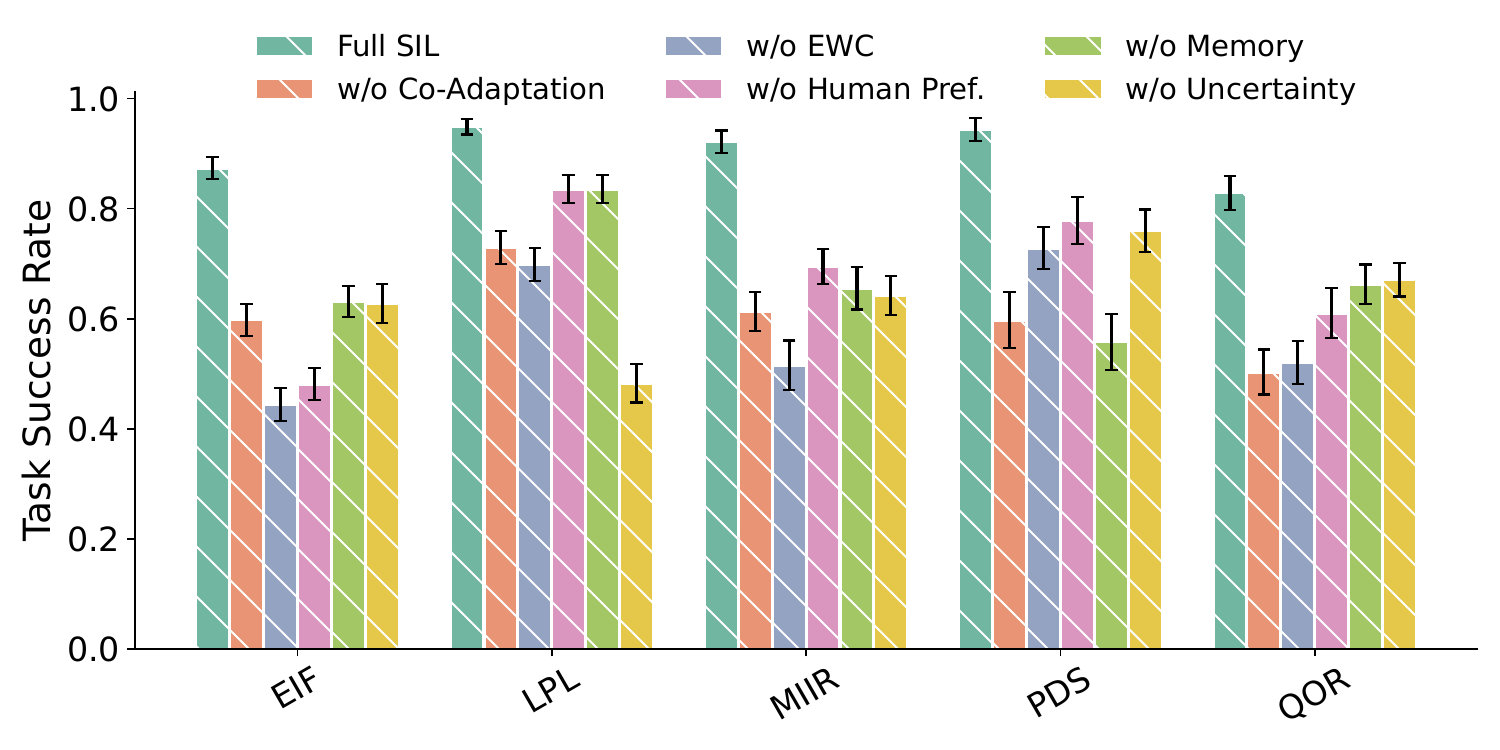}
        \caption{Task success rate across domains and ablated variants. Full SIL consistently outperforms all ablations, achieving near-ceiling performance on LPL, MIIR, and PDS. The worst performance arises when co-adaptation and EWC are disabled, confirming their critical role.
        }
        \label{fig:task-succes-rate}
        \vspace*{-1.10\baselineskip}
\end{figure}

    Table~\ref{tab:system_comparison} and Fig.~\ref{fig:task-succes-rate} report SIL’s performance across all task domains. SIL consistently outperforms all baselines on every metric. Most notably, it achieves a mean task completion rate of $87 - 94\%+$. This represents an absolute improvement of nearly $20$ points over the best ablation variants.   
\begin{table}[h!]
        \caption{ Performance comparison of SIL across task domains. Accuracies are averaged, and the stds are within $\pm 0.2$.}
        \centering
        \begin{tabular}{lccccc}
            \toprule
            \textbf{Metrics} & \textbf{EIF} & \textbf{MIIR} & \textbf{QOR} & \textbf{PDS} & \textbf{LPL} \\
            \midrule
            \textbf{TCR (\%)} $\uparrow$ & 87.36 & 92.18 & 82.88 & 94.35 & 94.89\\
            \textbf{CE} $\downarrow$ & 0.79 & 0.43 & 0.60 & 0.43 & 0.05 \\
            \textbf{BA ($\rho$)} $\uparrow$ & 0.76 & 0.86 & 0.80 & 0.84 & 0.89 \\
            \bottomrule
        \end{tabular}
        \label{tab:system_comparison}
        \vspace*{-.75\baselineskip}
\end{table}
\begin{figure}
        \centering
        \includegraphics[width=0.85\linewidth]{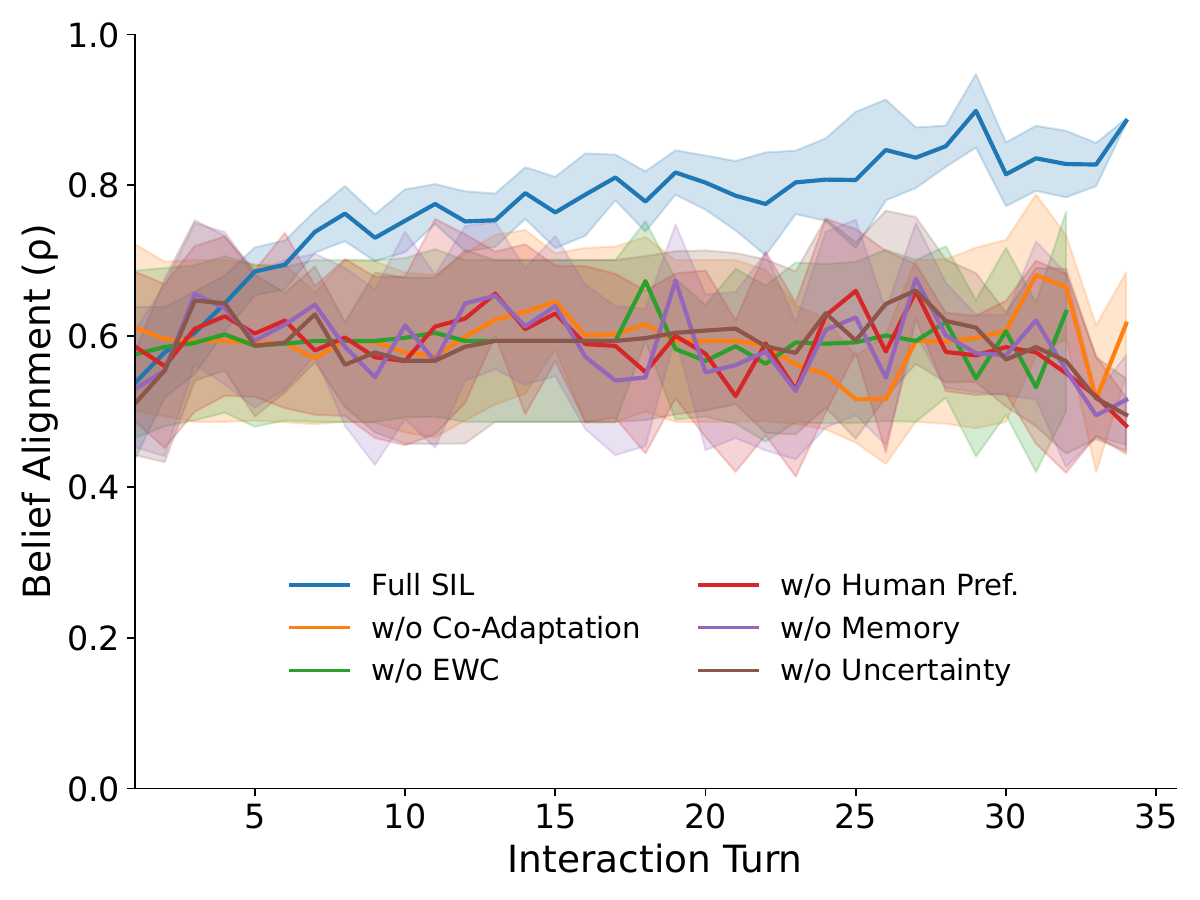}
        \caption{Mean belief alignment across the multi-turn interaction episodes (Section~\ref{sec:task-domain}). Full SIL exhibits rapid convergence toward stable equilibrium $\rho \approx 0.83$, maintaining high alignment throughout. Contrarily, ablation variants failed to achieve strong alignment $(\rho \approx 0.52 - 0.65)$.}
        \label{fig:alignment}
\end{figure}
     Figure~\ref{fig:alignment} demonstrates the core strength of SIL: bidirectional belief convergence. While the ablations show fluctuations around the suboptimal misalignment threshold, full SIL exhibits rapid convergence toward a stable equilibrium $\rho \approx 0.83$, with high belief alignment sustained throughout the multi-turn interaction.
    In contrast, ablations fluctuate around $\rho \approx 0.52 - 0.65$ and fail to achieve sustained alignment. From Table~\ref{tab:ablation}, the static LLM baseline achieves a TCR of only $60.1\%$, with no belief alignment and memory capabilities. 
    
\begin{figure*}
    \centering
    \includegraphics[width=0.95\linewidth]{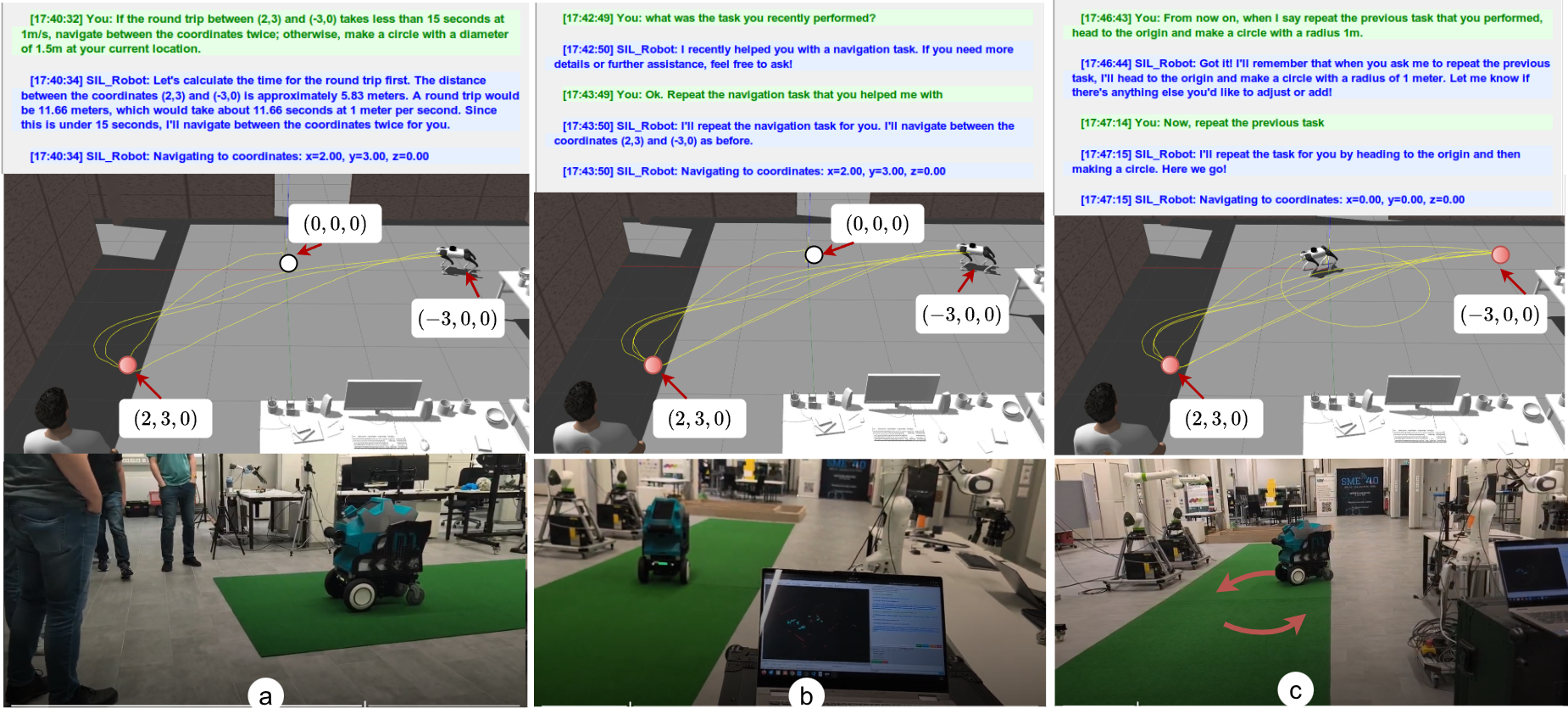}
    \caption{Qualitative examples of SIL in multi-turn interaction tasks. Yellow paths indicate the agent’s navigation trajectories, starting from the origin $(x=0,y=0,z=0)$.
    (a) The user issues a conditional navigation command requiring logical reasoning over spatial constraints; SIL computes the round-trip time and executes the correct policy.
    (b) The user probes anti-forgetting; SIL recalls and reproduces a previously executed navigation sequence, showing stable task memory.
    (c) The user teaches a new preference (``repeat previous task" implies returning to the origin and drawing a circle). SIL encodes this personalisation and applies it correctly in subsequent interactions, demonstrating preference retention and continual learning.}
    \label{fig:qualitative-visualisation}
    \vspace*{-1.15\baselineskip}
\end{figure*}

    Qualitatively, the static baseline fails in three characteristic ways: (i) it cannot resolve ambiguous references (e.g., `go there' produces either a refusal or a random guess), (ii) it cannot recall past interactions (e.g., `what did you do last?' yields a generic response), and (iii) it cannot retain learned preferences (e.g., taught aliases are forgotten immediately). CE and BA metrics are undefined for the static baseline as it lacks the corresponding mechanisms.
    Fig.~\ref{fig:qualitative-visualisation} shows representative qualitative results. These dialogues demonstrate SIL’s ability to combine logical reasoning, memory-based recall, and preference retention to sustain multi-turn interactions.

\begin{table}[h!]
    \centering
    \caption{Ablation study on SIL's core architecture. Accuracies are averaged, and the stds are within $\pm 0.2$.}
    \label{tab:ablation}
    \begin{tabular}{lccc}
    \toprule
    \textbf{Model Variant} & \textbf{TCR (\%) $\uparrow$} & \textbf{CE $\downarrow$} & \textbf{BA ($\rho$) $\uparrow$} \\
    \midrule
    Static LLM & 60.1 & --  & -- \\
    w/o Co-Adaptation & 61.7 & 0.55 & 0.52 \\
    w/o Episodic Memory & 68.3 & 0.50 & 0.59 \\
    w/o EWC Safeguard & 67.3 & 0.56 & 0.65 \\
    w/o Uncertainty Quantification & 62.8 & 0.52 & 0.55 \\
    w/o Human Preference Model & 66.0 & 0.51 & 0.59 \\
    \textbf{SIL (Full)} & \textbf{90.4} & \textbf{0.46} & \textbf{0.83} \\
    \bottomrule
    \end{tabular}
    \vspace*{-1.10\baselineskip}
\end{table}

\subsection{Ablation Study}\label{sec:ablation}
    We conducted ablation on SIL under five conditions to disentangle the contribution of each component. The results are shown in Table~\ref{tab:ablation}. The static GPT-4o~\cite{hurst2024gpt} LLM without memory or adaptation represents the traditional unidirectional baseline (master-apprentice configuration).
    Ablating the co-adaptation mechanism results in the largest performance drop, reducing the TCR to near-static-LLM levels $(61.7\%)$.
    Disabling EWC induced catastrophic forgetting, particularly evident on MIIR tasks, where the previously learned aliases are forgotten after distractor tasks. Memory, human preference modelling, and uncertainty contribute smaller but significant performance improvements, with the largest gains observed in context-intensive tasks (MIIR, QOR) and those requiring fine-grained personalisation (LPL).
    
\section{Conclusion}\label{sec:conclusion}
    In this paper, we address the master–apprentice challenge in natural language-conditioned human–robot interaction. We introduced SIL, a symbiotic interaction framework that enables co-adaptation between humans and agents within a shared latent task space.  We showed through empirical evaluation that unidirectional approaches, such as static LLM-based language-to-action pipelines, create unsustainable asymmetries across multi-turn interactions.
    In contrast to the ablated baselines, SIL achieved superior efficiency, yielding, on average, $0.46$ clarification requests per task, and a task completion rate of $90\%$. Moreover, belief alignment remained consistently high ($\rho \approx 0.83$) across the different task domains. Together, these results demonstrate that bidirectional co-adaptation enables robust and efficient collaboration over multi-turn interactions. Our future work will extend SIL to multi-user settings with conflicting or shifting preferences, alongside addressing its computational efficiency for deployment in resource-constrained robots.


\bibliographystyle{ieeetr}
\bibliography{references}



\section{APPENDIX}
\subsection{Encoder Architecture and Hyperparameters}\label{sec:imp-details}
Table~\ref{tab:impl-details} presents the encoder architecture employed in SIL, and summarises the key hyperparameters used across all experiments.
The encoder input embeddings are produced by a frozen \texttt{all-mpnet-base-v2} sentence transformer.
The memory retrieval pipeline (Section~\ref{sec:mem}) uses a separate \texttt{paraphrase-MiniLM-L6-v2} model for computing semantic similarity scores.
The encoder is optimised with Adam ($lr = 0.001$).
For more details, we refer the reader to:~\url{https://linusnep.github.io/SIL/}.
\begin{table}[ht]
\caption{Encoder architecture and key hyperparameters used in SIL.}
\centering
\setlength{\tabcolsep}{1pt}
\renewcommand{\arraystretch}{0.90}
\begin{tabular}{ccccc}
\toprule
\multicolumn{5}{c}{\textbf{Encoder Architecture}}\\
\midrule
\textbf{Layer} & \textbf{Input} & \textbf{Output} & \textbf{Activation} & \textbf{Other}\\
\midrule
1 & 768 & 512 & ReLU & BN, Drop(0.2)\\
2 & 512 & 384 & ReLU & BN\\
3 & 384 & $d{=}256$ & Tanh & -- \\
\midrule
\multicolumn{5}{c}{\textbf{Hyperparameters}}\\
\midrule
\multirow{3}{*}{\rotatebox{0}{Belief update}}
 & Latent dim.\ $d$ & \multicolumn{3}{c}{256}\\
 & $\eta_{1{-}6}$ & \multicolumn{3}{c}{0.6, 0.3, 0.1, 0.7, 0.2, 0.1}\\
 & $\alpha^A,\alpha^H, \tau_{\text{mis}}$ & \multicolumn{3}{c}{0.1,\;0.05,\;0.6}\\
\midrule
\multirow{2}{*}{\rotatebox{0}{Initialisation}}
 & $\mathbf{k}^H_0,\mathbf{k}^A_0$ & \multicolumn{3}{c}{0.8,\;0.5}\\
 & $\sigma_{\text{init}}$ & \multicolumn{3}{c}{0.1}\\
\midrule
\multirow{2}{*}{\rotatebox{0}{Triplet / EWC}}
 & Margin $m$/ EWC $\lambda$ & \multicolumn{3}{c}{1.0\;/\;2000}\\
 & Pos./neg.\ sim.\ thresh. & \multicolumn{3}{c}{$>0.7$ / $<0.3$}\\
\midrule
\multirow{2}{*}{\rotatebox{0}{Memory}}
 & Buffer capacity $\mathcal{M}_{\text{ep}}$ & \multicolumn{3}{c}{2000}\\
 & Retrieval weights $w_s,w_b$ & \multicolumn{3}{c}{0.6,\;0.4}\\
\bottomrule
\end{tabular}
\label{tab:impl-details}
\vspace*{-1.10\baselineskip}
\end{table}



\end{document}